\documentclass{sig-alternate-05-2015}

\usepackage{CJKutf8} 
\usepackage{ucs}
\usepackage[encapsulated]{CJK} 
\newcommand{\myfont}{gkai} 
\usepackage{latexsym}
\usepackage{array}
\usepackage{multirow}
\usepackage{amsmath}
\usepackage{amsfonts}
\usepackage{graphicx}
\usepackage{color}
\usepackage{algorithm}
\usepackage{mathrsfs}
\usepackage{algpseudocode}
\usepackage{enumitem}
\usepackage{arydshln}
\definecolor{Orange}{rgb}{1,0.5,0}

\setitemize[1]{itemsep=0pt,partopsep=0pt,parsep=\parskip,topsep=0pt}
\setenumerate[1]{itemsep=0pt,partopsep=0pt,parsep=\parskip,topsep=0pt}

\date{}

\begin{document}

\begin{CJK}{UTF8}{\myfont}
\doi{10.475/123_4}

\isbn{123-4567-24-567/08/06}


\acmPrice{\$15.00}

\setcopyright{acmcopyright}

\title{Aligning Coordinated Text Streams through Burst Information Network Construction and Decipherment}

\numberofauthors{6} 
\author{
\alignauthor
Tao Ge\\
       \affaddr{EECS, Peking University}\\
       \affaddr{Beijing, China}\\
       \email{getao@pku.edu.cn}
\alignauthor
Qing Dou\\
       \affaddr{Facebook, Inc.}\\
       \affaddr{Menlo Park, CA, USA}\\
       \email{qdou@isi.edu}      
\alignauthor
Xiaoman Pan\\
       \affaddr{Rensselaer Polytechnic Institute}\\
       \affaddr{Troy, NY, USA}\\
       \email{panx2@rpi.edu}
\and
\alignauthor
Heng Ji\\
       \affaddr{Rensselaer Polytechnic Institute}\\
       \affaddr{Troy, NY, USA}\\
       \email{jih@rpi.edu}      
\alignauthor
Lei Cui\\
      \affaddr{Microsoft Research}\\
     \affaddr{Beijing, China}\\
    \email{lecu@microsoft.com} 
    \alignauthor
Baobao Chang\\
      \affaddr{EECS, Peking University}\\
     \affaddr{Beijing, China}\\
    \email{chbb@pku.edu.cn} 
    \and
    \alignauthor
Zhifang Sui\\
      \affaddr{EECS, Peking University}\\
     \affaddr{Beijing, China}\\
    \email{szf@pku.edu.cn}      
        \alignauthor
Ming Zhou\\
      \affaddr{Microsoft Research}\\
     \affaddr{Beijing, China}\\
    \email{mingzhou@microsoft.com}      
}

\maketitle
\begin{abstract}



Aligning coordinated text streams from multiple sources and multiple languages has opened many new research venues on cross-lingual knowledge discovery. In this paper we aim to advance state-of-the-art by: (1). extending coarse-grained topic-level knowledge mining to fine-grained information units such as entities and events; (2). following a novel ``\emph{Data-to-Network-to-Knowledge (D2N2K)}'' paradigm to construct and utilize network structures to capture and propagate reliable evidence.  We introduce a novel Burst Information Network (BINet) representation that can display the most important information and illustrate the connections among bursty entities, events and keywords in the corpus. We propose an effective approach to construct and decipher BINets, incorporating novel criteria based on multi-dimensional clues from pronunciation, translation, burst, neighbor and graph topological structure. The experimental results on Chinese and English coordinated text streams show that our approach can accurately decipher the nodes with high confidence in the BINets and that the algorithm can be efficiently run in parallel, which makes it possible to apply it to huge amounts of streaming data for never-ending language and information decipherment.

\vspace{-0.1cm}

\vspace{-0.1cm}



\end{abstract}

\section{Introduction}\label{sec:intro}


In this information era, a large amount of text data is continuously produced over time in various applications (e.g., news agencies and social networks), which results in massive streams of text data with much valuable information and knowledge. Streaming data is an important source for many applications such as disaster summarization~\cite{Kedzie2015}.
There are many studies on pattern mining from text data streams but most of them focus on a single text stream. Nonetheless, there are many text streams that are topically related and indexed by the same
set of time points, which are called \emph{coordinated text streams}~\cite{Wang2007}. For example, news streams from a news agency in different languages are coordinated. In recent, text mining on coordinated text streams has gained attention in both data mining and natural language processing communities because one can discovery useful facts and important patterns by aligning the coordinated streams. 
For example, \cite{Wang2007} and \cite{zheng2012cross} tried to discover common topic patterns from coordinated text streams. In this paper, we aim to advance state-of-the-art from the following two perspectives.

(1) \textbf{Aligning Fine-grained Information Units instead of Coarse-grained Topics}.

\begin{figure}[t]
\centering
\includegraphics[width=8cm]{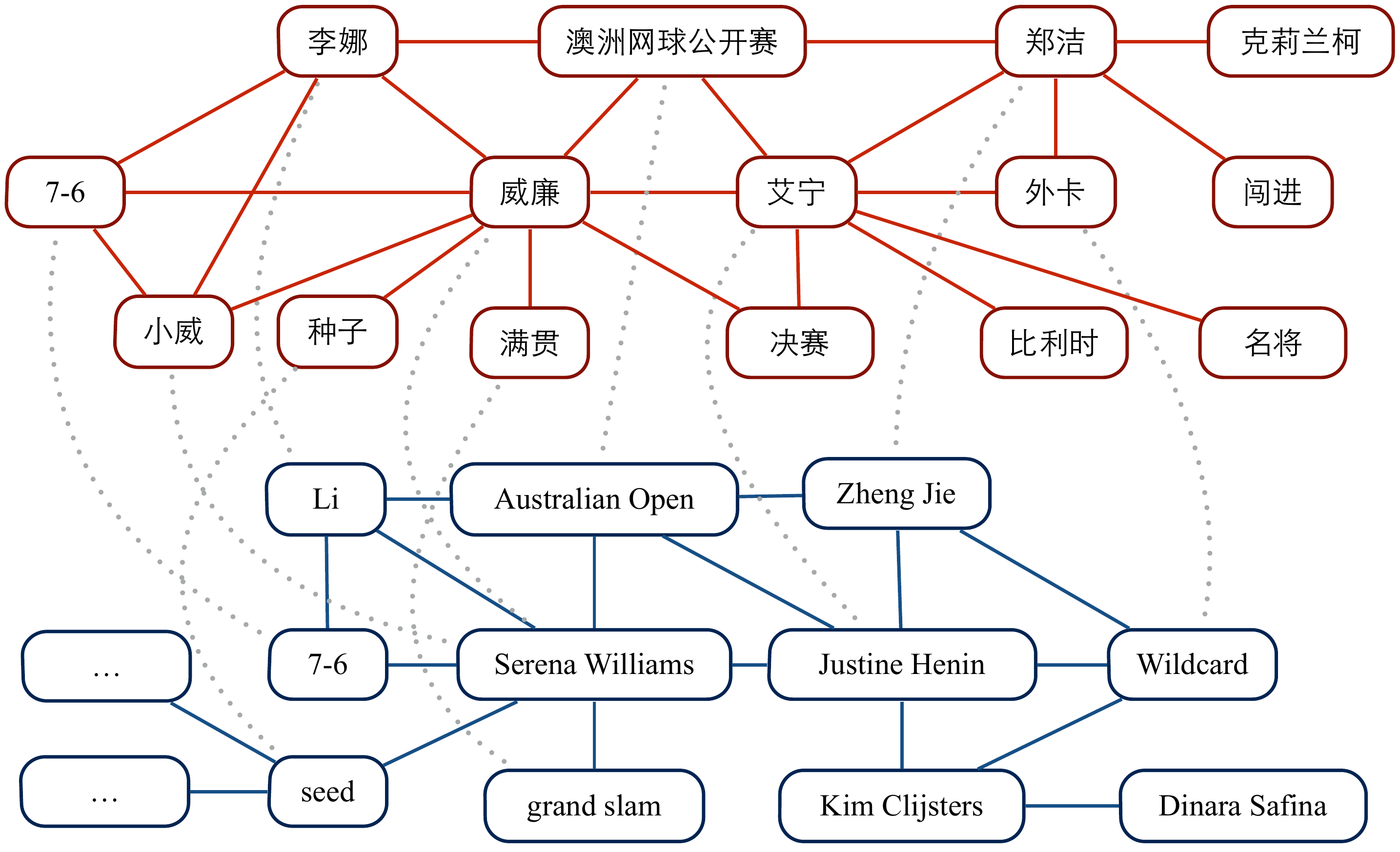}
\caption{(A small part of) Burst information networks constructed from Chinese and English. Due to the space limitation, we do not present the information of the burst periods of nodes in this figure.
\label{fig:walkthrough}}\vspace{-1.0cm}
\end{figure} 

Most documents in coordinated streams are not entirely parallel. Fortunately, topically related and temporally correlated important facts such as bursty events and entities often repeatedly co-occur across streams from different languages.  
Most previous work on coordinated text stream mining such as~\cite{Wang2007} analyzed the streams on the coarse-grained topic level, which inevitably leads to loss of some information. For instance, a topic-level cross-lingual text stream analysis cannot inform us the mentions of an entity in the other text stream. For fully exploiting the coordinated text streams, we propose to analyze the text streams by attempting to align important fine-grained information units (e.g., entities and events) across the streams. Specifically, in this paper, we focus on aligning cross-lingual coordinated text streams. We choose two very different languages -- Chinese and English streams for alignment as a case study. Aligning cross-lingual coordinated text streams is a meaningful task because it can mine translation pairs for completing a bi-lingual lexicon, which would be helpful for cross-lingual information extraction, information retrieval and machine translation systems that suffer from bi-lingual lexicons with a limited number of entries.


(2) \textbf{Data-to-Network-to-Knowledge (D2N2K) instead of Data-to-Knowledge (D2K)}

Instead of directly turning massive, unstructured data streams into structured knowledge (D2K), we propose a new Data-to-Network-to-Knowledge (D2N2K) paradigm, based on the following observations: (i) most information units are not independent, instead they are interconnected or interacting, forming massive networks; (ii) if  information networks can be constructed across multiple languages, they may bring tremendous power to make knowlege mining algorithms more scalable and effective because we can take advantage of the graph structures to acquire and propagate knowledge. For aligning the cross-lingual text streams, we first propose a novel text stream's structured representation called ``\emph{Burst Information Networks (BINets)}'' to focus on the most important information units. In BINets, nodes are bursty words including important entities and events, and edges represent their co-burst relations. An example is depicted in Figure~\ref{fig:walkthrough}. Similar to traditional Information Networks~\cite{Han2010,Li2014}, nodes within a community in BINets are semantically related (e.g., entities are often participants in certain events). Moreover, they are also temporally coherent. For example, the nodes in Figure~\ref{fig:walkthrough} are all related to the 2010 Australian Open.






After the BINets are constructed, our next goal is to align the nodes in BINets across text streams. In comparable corpora, bursty words which are about either a breaking news or a hot topic usually burst across languages. For example, both ``澳洲网球公开赛(Australian Open)'' and its translation in English ``Australian Open'' in figure \ref{fig:walkthrough} burst from Jan 14 to Jan 31 in 2010. We take advantage of such characteristics of bursts across languages and propose four decipherment clues from some starting points (prior knowledge) in a propagation manner. Experimental results demonstrate that our aligning approach can accurately mine cross-lingual translation pairs and that the more data provided, the more translation pairs will be discovered, which means that if our approach can run over the streaming data produced throughout the time, it will be a never-ending translation pair discovery framework.






\section{Burst Information Network}
We will start by defining some new terminologies and describing the detailed methods for constructing Burst Information Networks.
\subsection{Burst Detection}
\begin{figure}[t]
\centering
\includegraphics[width=8.1cm]{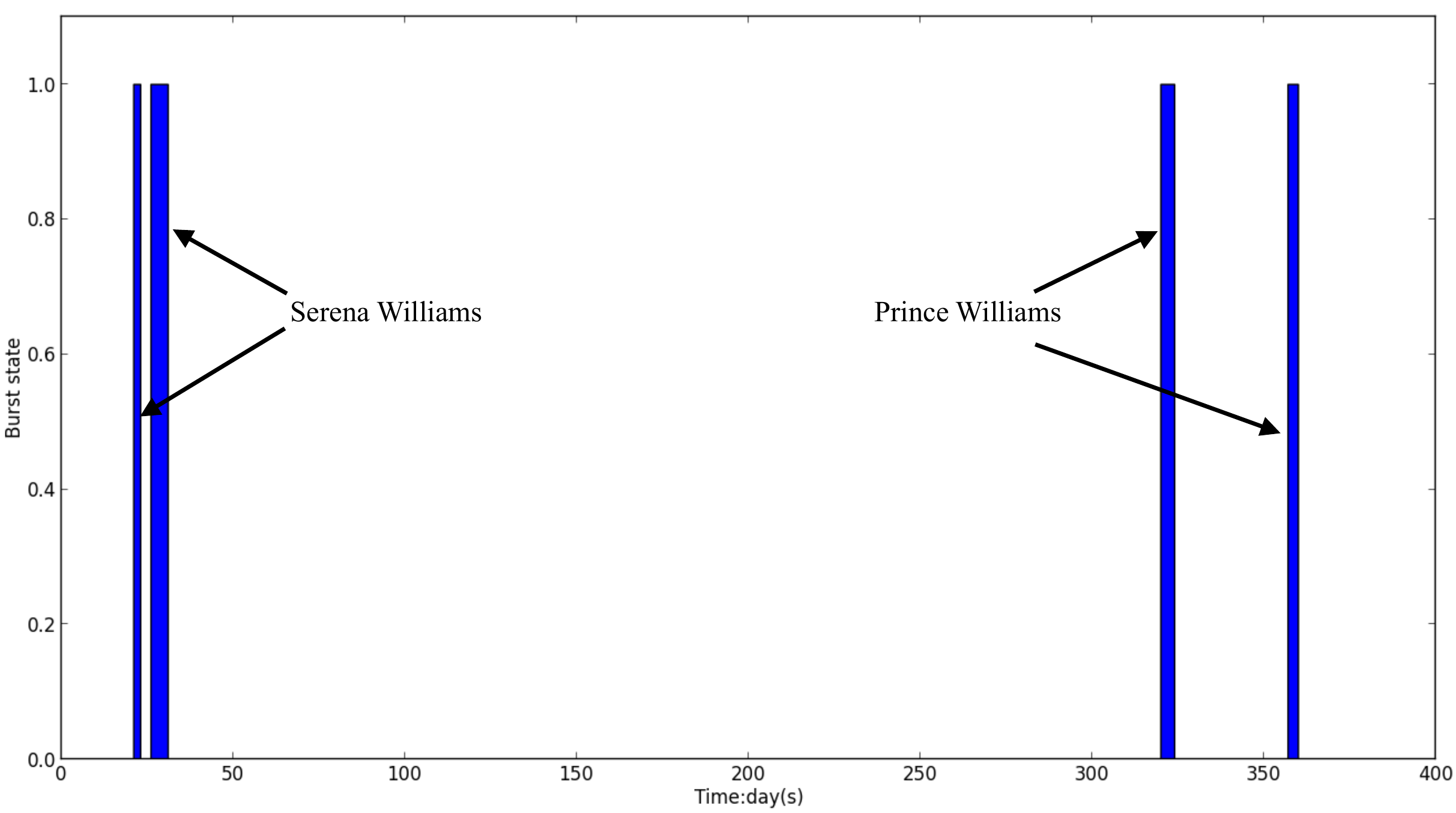}
\caption{Burst state of ``威廉(Williams)'' during 2010. The horizontal axis is time (days after Jan 1, 2010) and the vertical axis is the burst state of the word, which is either 1 or 0 to indicate if the word bursts or not. The first two bursts refer to Serena Williams while the last two bursts refer to Prince Williams.\label{fig:bstate}}\vspace{-0.4cm}
\end{figure}

A word's burst refers to a remarkable increase in the number of occurrences of the word during a short period and it might indicate important events or trending topics. For example, the word ``威廉(Williams)'' has a burst from Jan 27 to Jan 31, as shown in Figure \ref{fig:bstate} because of the wonderful performance of Serena Williams in the 2010 Australian Open held during this period. For a timestamped document collection $\mathbf{C}=\{D_1,D_2,...,D_\tau,...,D_{T}\}$ where $D_\tau$ is the collection of documents during time $\tau$, we define a word $w$'s burst sequence $\boldsymbol{s}=(s_1,s_2,...,s_\tau,...,s_T)$ in which $s_\tau$ is either 1 or 0 to indicate if $w$ bursts or not at the time epoch. Most burst detection approaches (e.g., \cite{kleinberg2003bursty}) detect a word's burst states mainly based on comparing a word's probability $q^t$ at a time epoch $t$ to its base probablity $q_0$ that is the word's global probability throughout the stream. In this paper, we use the approach of \cite{zhao2012novel} which is a variant of \cite{kleinberg2003bursty} for burst detection.

Specifically, if a word $w$ bursts at every time $\tau$ during a period, we call this period as a burst period of $w$, and $w$ has a burst during this period. In Figure \ref{fig:bstate}, we can see that the word ``威廉(Williams)'' has 4 burst periods. 
Formally, we use $\mathcal{P}$ to denote a burst period $w$. $\mathcal{P}$ is defined as a consecutive time sequence during which the word bursts at every time epoch $\tau$:

\centerline{
$\mathcal{P}=(\tau_i,\tau_{i+1},\tau_{i+2},...,\tau_{i+n})$
}
\centerline{
$\forall \tau \in \mathcal{P}~~s_\tau=1$
}
\noindent where $s_\tau$ denotes the burst state at time $\tau$.
\subsection{Burst Information Network Construction}
An information network is a graph depicting connections between entities and events, in which every community should be topically and temporally coherent, as shown in Figure \ref{fig:walkthrough}. Most of the previous approaches (e.g., ~\cite{Li2014}) relied on a large number of linguistic resources and labeled data to construct information networks.
\cite{sayyadi2013graph} proposed a co-occurrence based approach to build an information network where each node is a word type (i.e., term) and edges between nodes denote the co-occurrence of words in a document. The resulting information network can demonstrate the connections between words. It usually consists of several communities (i.e., a group of nodes that are strongly related) and each community in the information network corresponds to a topic. However, such kind of information networks did not incorporate temporal information and thus the communities are usually not temporally coherent. In addition, it did not tackle the ambiguity problem since a word may have different senses in the corpus. To address these problems, we propose a new representation called ``\emph{Burst Information Network (BINet)}" by using burst elements as nodes:

\textbf{A Burst Element} is a burst of a word. It can be represented by a tuple: $\langle w,\mathcal{P} \rangle$ where $w$ denotes the word type and $\mathcal{P}$ denotes one burst period of $w$.

A word type may have multiple burst periods while a burst element only has one burst period. A word during its different burst periods will be regarded as different burst elements. For example, in Figure \ref{fig:bstate}, ``威廉(Williams)'' has 4 bursts, each of which is regarded as a distinct burst element.

In general, there are two main advantages using burst elements as nodes to build the information network:

\begin{itemize}
\item A burst element not only includes semantic information but also incorporates the temporal dimension. It cannot co-occur with another burst element that does not burst during the same burst period. Therefore, the communities in this information network must be both topically and temporally coherent.

\item Since a burst element denotes a bursty word during a consecutive period, its sense is likely to be consistent. Multiple bursts of a word will be considered as different burst elements. Therefore, each node in the information network is less ambiguous. For example, even though ``威廉(Williams)'' (shown in Figure \ref{fig:bstate}) refers to different entities in the corpus, its sense in a single burst is consistent. It refers to ``Serena Williams'' in the first two bursts, while it refers to ``Prince Williams'' in the last two bursts.
\end{itemize}

Formally, we define the BINet $G= \langle V,E \rangle$ as follows. Each node $v \in V$ is a burst element and each edge $\epsilon \in E$ denotes the correlation between burst elements. Intuitively, if two burst elements frequently co-occur in a document, then they are strongly related and thus the edge between these two nodes should be highly weighted. Therefore, we define the weight $\omega$ of an edge between $v_i$ and $v_j$ as the number of documents $v_i$ and $v_j$ co-occur.

\vspace{-0.2cm}\section{Decipherment}\vspace{-0.2cm}
\subsection{Hypothesis and Overall Framework}
After constructing BINets from a foreign language, we aim to decipher them into English by translating as many nodes as possible. In this paper we use Chinese as a foreign language and conduct experiments with as few language-specific resources as possible. Our basic hypothesis is that co-burst information units  (entities, events, etc.) tend to co-occur across many different languages. Therefore, if we gather a collection of English documents created from a similar time frame as the Chinese corpus, we can construct BINets from this English corpus and then consult them to validate candidate translations. Formally, we define $G_c=\langle V_c,E_c \rangle$ and $G_e= \langle V_e,E_e \rangle$ as the Chinese BINet and English BINet respectively. For those who do not know Chinese, $G_c$ can be considered as a network of ciphers. 
Based on this hypothesis we design a novel BINet decipherment procedure to use $G_e$ to decipher $G_c$. Formally, we define the decipherment as a process to find a counterpart $e \in V_e$ of a node $c \in V_c$ so that $e$ is $c$'s translation or $c$ is a mention referring to the entity $e$ refers to.\footnote{$c$ and $e$ are burst elements (i.e., nodes in the BINets) defined in Section 2. Sometimes, we also use $c$ and $e$ to denote the nodes' words if that does not lead to misunderstanding.}









\vspace{-0.1cm}\subsection{Starting Point}


To start deciphering the Chinese BINet, we need a few seeds based on prior knowledge as a starting point. Inspired by some previous work on bi-lingual dictionary induction (e.g.,~\cite{koehn2002learning,Irvine2013}), decipherment (e.g.,~\cite{Ravi2011,Dou2012,Dou2014}) and name translation mining (e.g.,~\cite{ji2009mining,Lin2011}), we utilize a few linguistic resources - a bi-lingual lexicon and language-universal representations such as time/calendar date, number, website URL, currency and emoticons to decipher a subset of Chinese nodes. For the example shown in Figure \ref{fig:walkthrough}, we can decipher some nodes in the Chinese BINet such as ``7-6'' (to ``7-6'') and ``种子'' (to ``seed'').



\subsection{Candidate Generation}\label{subsec:candidate}

Based on the prior knowledge mentioned above, we can decipher a subset of nodes in the Chinese BINets. 
For the remaining nodes, we first need to discover its possible candidate translations.
For a node $c$ in the Chinese BINet, its translation $e$ can be any node in the English BINet or does not exist in the English BINet. However, if we use all nodes in the English BINet as its candidates, the candidate list will be extremely long, which makes it inefficient to validate each candidate. To address this problem, we exploit temporal constraints to dramatically narrow down the search scope.

As defined in Section 2, a node in a BINet is a burst element that has a burst period. For a node in the Chinese BINet, its candidate translation is likely to be a node with the same burst period in the English BINet. For example, the node ``威廉(Williams)'' in the Chinese BINet shown in Figure \ref{fig:walkthrough} bursts from Jan 27 to Jan 31, 2010. Thus we should look for its translation from the nodes in the English BINet whose burst period overlaps with this period. The reason is that a burst of a word usually indicates an important event which is usually reported by multiple news agencies in multiple languages. 
Therefore, for a node $c \in V_c$ in the Chinese BINet, its candidate translations can be derived as follows:
\begin{equation}\nonumber
Cand(c) = \{e|\mathcal{P}(e) \cap \mathcal{P}(c) \neq \emptyset \}
\end{equation}
where $e \in V_e$, and $\mathcal{P}(c)$ and $\mathcal{P}(e)$ are the burst periods of $c$ and $e$ respectively.



\vspace{-0.1cm}\subsection{Candidate Validation}\label{subsec:validation}


After we obtain the candidate list of $c$ (i.e., $Cand(c)$), we need to validate each node $e \in Cand(c)$ and choose the most possible one as the translation of $c$ so that we can decipher the BINet. 
Formally, we define the credibility score $Score(c,e)$ as the credibility score of $e$ being the correct translation of $c$. We propose the following novel criteria for validation.

\subsubsection*{Pronunciation}
For a node $e \in Cand(c)$, if its pronunciation is similar to $c$, then $e$ is likely to be the translation of $c$. For example, the pronunciation of ``Williams'' is similar to ``威廉 (pronounced as Wei Lian)'' in Chinese, which is an important clue to infer that ``Serena Williams'' is likely to be the translation of ``威廉(Williams)''.

We compute the normalized edit distance (i.e., Levenshtein distance) between Chinese pinyin\footnote{pinyin is the official romanization system for Chinese. (https://en.wikipedia.org/wiki/Pinyin)} of $c$ and English candidate $e$ as the pronunciation similarity: 
\begin{equation}\nonumber
\small
Leven' = \frac{Leven(pinyin(c),e)}{len(e)}
\end{equation}
where $len(e)$ is the length of the string of $e$. For a multi-word candidate, we compute the normalized Levenshtein distance for each part and use the minimal value.

We define $S_p$ as a score to reflect the credibility of a candidate being the correct translation based on the pronunciation similarity:

\begin{equation*}
\small
S_p = 
\begin{cases}
3 & \text{if}~0\le Leven' < 0.25, \\
1 + \frac{1}{2Leven'} & \text{if}~0.25<Leven' \le 0.5 ,\\
0 & \text{else}
\end{cases}
\end{equation*}

\subsubsection*{Translation}
For a node $e \in Cand(c)$, it is possible that it or some part of it exists in the bi-lingual lexicon. We can exploit the translation clue to validate if $e$ is the translation of $c$. For example, ``Australian Open'' is a candidate of ``澳洲网球公开赛(Australian Open)'' 
as shown in figure \ref{fig:walkthrough}. Even though ``澳洲网球公开赛(Australian Open)'' is not in the bi-lingual lexicon, ``Australian'' and ``open'' are in the lexicon and their Chinese translations are ``澳洲的(Australian)'' and ``公开赛(open)'' respectively. Since ``澳洲的(Australian)'' and ``公开赛(open)'' have long common subsequences with ``澳洲网球公开赛(Australian Open)'', we can infer that ``Australian Open'' is likely to be the appropriate translation of ``澳洲网球公开赛(Australian Open)''.

We first extract all Chinese translations $C(e)$ of the word type of $e$ from the bi-lingual lexicon. 
If the word type of $e$ is a multi-word, we concatenate translations of its parts. We define a score $S_t$ to measure the credibility of a candidate being the correct translation based on the length of the longest common subsequence of $c$ and translation of $e$:

\begin{equation*}
\small
S_t =
\begin{cases}
2 & \text{if}~\frac{LCS(c,c^*)}{len(c)}\ge0.75,\\
\frac{1}{2(1-\frac{LCS(c,c^*)}{len(c)})} & \text{if }0.5\le\frac{LCS(c,c^*)}{len(c)}<0.75, \\
0 & \text{else}
\end{cases}
\end{equation*}
where $c^*=\arg \max \limits_{c' \in C(e)} LCS(c,c')$, $LCS(c,c^*)$ denotes the length of the longest common sequence between string $c$ and $c^*$, and $len(c)$ is the length of string of $c$.

\subsubsection*{Neighbor}


The graph topological structure of each BINet is also an important clue for decipherment. By analyzing a node's neighbor, we can learn useful knowledge to decipher the node. For the example in Figure \ref{fig:walkthrough}, ``艾宁(Henin)''\footnote{``艾宁'' is the translation of ``Justine Henin'' in the AFP Chinese corpus we used, while ``Justine Henin'' is more commonly translated to ``海宁'' in other Chinese corpora.} in the Chinese BINet has neighbors such as ``威廉(Williams)'', ``澳洲网球公开赛(Australian Open)'' and ``外卡(wildcard)'' while ``Justine Henin'' in the English BINet is connected with ``Serena Williams'', ``Australian Open'' and ``wildcard''. If we know ``Serena Williams'', ``Australian Open'' and ``wildcard'' are the translations of `威廉'', ``澳洲网球公开赛'' and ``外卡'' respectively, then we can infer that ``Justine Henin'' is likely to be the translation of ``艾宁''.

We define $N(c)$ and $N(e)$ to denote the set of adjacent nodes of $c$ in the Chinese BINet and the adjacent nodes of $e$ in the English BINet respectively. The neighbor validation score $S_n$ of $\langle c,e \rangle$ is defined as follows:

\begin{equation}
\small
S_n=\sum_{c' \in N(c)}\hat{\omega}_{c,c'}\max \limits_{e' \in N(e)}Score(c',e')
\end{equation}
where $Score(c',e')$ is the overall score of $e'$ being the translation of $c'$, as defined at the beginning of Section \ref{subsec:validation}, $\hat{\omega}_{c,c'}=\frac{\omega_{c,c'}}{\sum \limits_{c'' \in N(c')}\omega_{c',c''}}$, which is the normalized weight of the edge between $c$ and $c'$.




\subsubsection*{Co-burst}

\begin{figure}[t]
\centering
\includegraphics[width=6.5cm,height=7.5cm]{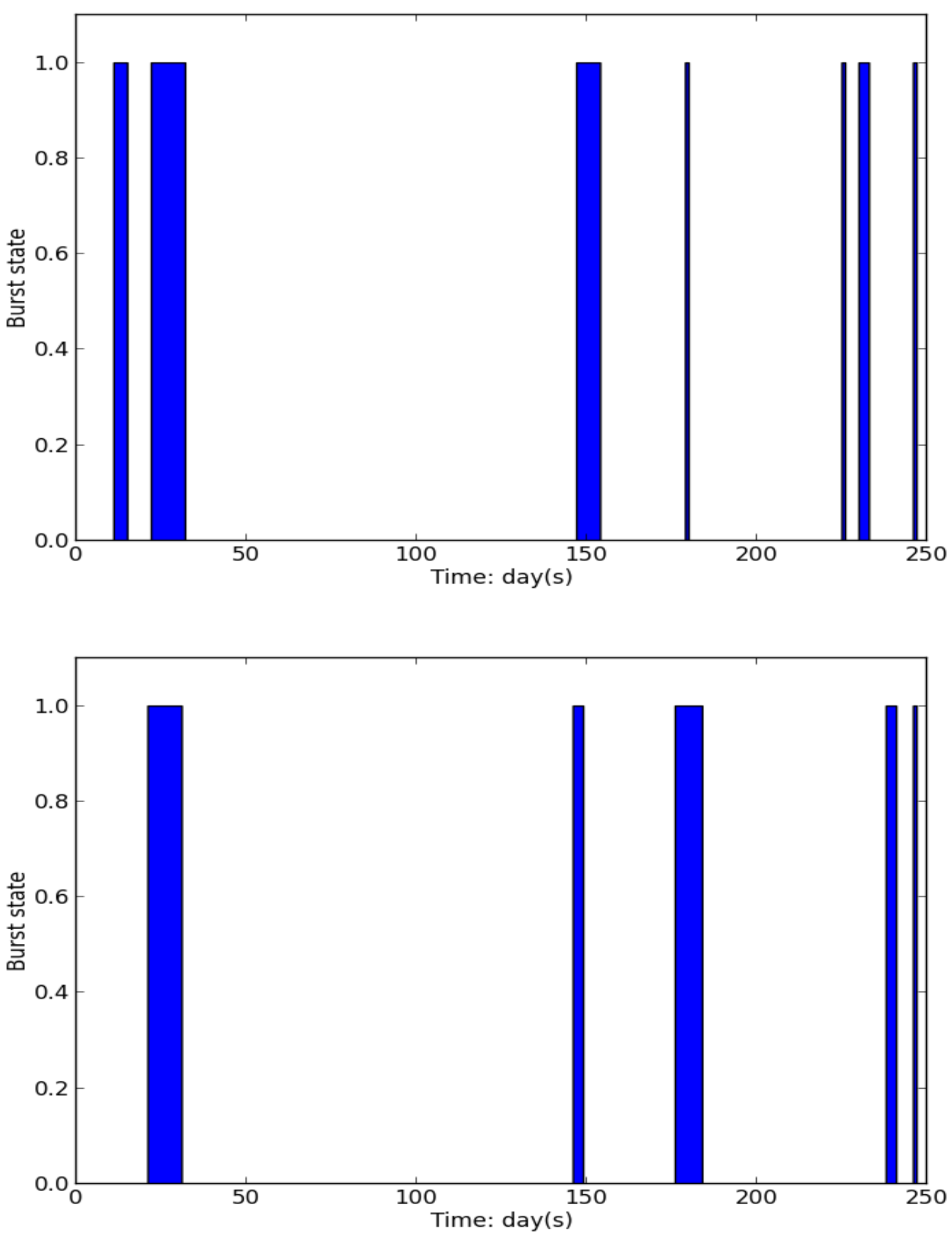}
\caption{Burst state of ``小威'' (the upper) and ``Serena Williams'' (the lower). Their bursts are correlated.\label{fig:coburst}}\vspace{-0.4cm}
\end{figure}

If the word type of $e \in Cand(c)$ frequently co-bursts with the word type of $c$, then $e$ is likely to be the translation of $c$. For example, ``Serena Williams'' in the English corpus usually co-bursts with ``小威'' in the Chinese corpus, as shown in figure \ref{fig:coburst}, which is a powerful clue to infer that ``Serena Williams'' is the translation of ``小威''.

We define $S_b$ as co-burst validation score as follows:

\begin{equation}\nonumber
\small
S_b = \frac{\boldsymbol{s_{w(c)}} \cdot \boldsymbol{s_{w(e)}}}{|\boldsymbol{s_{w(c)}|}^2+|\boldsymbol{s_{w(e)}}|^2}
\end{equation}
where $w(c)$ denotes the word type of the node $c$ and $\boldsymbol{s}_w$ denotes the burst sequence of the word $w$, as defined in Section 2. Note that in the above equation, we regard $\boldsymbol{s_{w(e)}}$ as a vector. The numerator is the number of days when $c$ and $e$ co-burst and the nominator is the sum of the number of days when $c$ and $e$ burst.

\subsection{Graph-based Decipherment}
Based on the clues introduced above, we define the overall validation (credibility) score by the following equation:

\vspace{-0.1cm}\begin{equation}
\small
Score(c,e) = f(S_p,S_t,S_n,S_b)
\end{equation}
where $f(\cdot)$ is a function and $S_p$, $S_t$, $S_n$ and $S_b$ are validation scores based on the pronunciation, translation, neighbor and co-burst clues respectively.
Specifically, 
we define:
\begin{equation}\label{eq:overallscore}
\small
\begin{split}
f(S_p,&S_t,S_n,S_b)=\\
&\eta S_p + \lambda S_t + \min(\gamma S_n, \hat{S}_n^{max}) + \delta S_b
\end{split}
\end{equation}
where $\eta$, $\lambda$, $\gamma$ and $\delta$ are parameters for adjusting weights of the above clues. Note that in Eq (\ref{eq:overallscore}), we constrain the effect of the neighbor validation score (no higher than $\hat{S}_n^{max}$). The reason is that the neighbor validation score $S_n$ will be interacted by other candidate pairs (i.e., $S_n$ depends on other candidate pairs' scores and it will influence other candidate pairs' score in turn) and it is unbounded because its definition is recursive. To limit the effect on the score and avoid error propagation, we constrain the effect of $S_n$ (i.e., the upper bound of $\gamma S_n$ is $\hat{S}_n^{max}$) on the overall score.

Based on Eq (\ref{eq:overallscore}), we can now compute the score of any candidate pair $\langle c,e \rangle$. For the pairs that are known to be correct alignments according to the prior knowledge, their scores will be fixed to 1. For other possible candidate pairs, we initialize their scores as follows:
\begin{equation}\label{eq:init}
\small
Score(c,e) = \frac{0.5}{|Cand(c)|}
\end{equation}
where $Cand(c)$ is the set of English translation candidate nodes and it is generated by the method mentioned in Section \ref{subsec:candidate}.

\begin{algorithm}[htbp]
\centering
\small
\caption{Graph-based Decipherment\label{alg:pagerank}}
\begin{algorithmic}[1]
\State For the determined pair $ \langle c,e \rangle$ based on the prior knowledge, $Score(c,e) \gets 1$ 
\State For other undermined pairs $\langle c,e \rangle$, initialize $Score(c,e)$ according to Eq (\ref{eq:init});
\For{each $iteration$ (until convergence)}
\For{each undetermined pair $\langle c,e \rangle$}
\State $new\_score \gets f(S_p,S_t,S_n,S_b)$ according to Eq (\ref{eq:overallscore});
\State $update(c,e) = \min(0.99,new\_score)$
\EndFor
\For{each undetermined pair $\langle c,e \rangle$}
\State $Score(c,e) \gets update(c,e)$
\EndFor
\EndFor
\end{algorithmic}
\end{algorithm}

Since the score of a pair $ \langle c,e \rangle$ is affected by the scores of other pairs, 
we use a label propagation algorithm to iteratively compute and update the scores so that we can decipher the entire Chinese BINet in a propagation manner. This procedure is elaborated in Algorithm \ref{alg:pagerank}. Note that for a candidate pair $\langle c$,$e \rangle$, the upper bound of its score is set to 0.99 to distinguish from the known alignments whose score is fixed to 1.

\vspace{-0.15cm}\section{Experiments}\vspace{-0.15cm}
\subsection{Data}

 
We use the 2010 Agence France Presse news story articles in Chinese Gigaword \cite{graff2005chinese} and English Gigaword \cite{graff2003english} as our non-parallel news stream corpora. 
The Chinese corpus has 17,327 documents and the English corpus contains 186,737 documents. We removed stopwords, conducted lemmatization and name tagging for the English corpus and did word segmentation and name tagging for the Chinese corpus using the Stanford CoreNLP toolkit~\cite{manning-EtAl:2014:P14-5}. Note that, we performed name tagging in the Chinese corpus just for avoiding segmenting a named entity to separate words and we did not use any other information (e.g., named entity type) from name tagging results because we want to prove that our approach is language-independent and does not need much language-specific knowledge. Therefore, in addition to a seed bilingual lexicon, a pinyin (pronunciation) dictionary and toolkits for Chinese word segmentation, we did not use any other Chinese-related resources in our experiments.

We detect bursts and construct the BINets for the Chinese and English stream respectively. 
The constructed Chinese BINet has 7,360 nodes and 33,892 edges while the English BINet has 8,852 nodes and 85,125 edges. The seed bi-lingual lexicon we used is released by \cite{zens2004improvements} which contains 81,990 Chinese word entries, each of which has an English translation. Among those 7,360 nodes in the Chinese BINet, there are 2,242 nodes that need to be deciphered because their words are not in the bi-lingual lexicon. 



\vspace{-0.1cm}\subsection{Experimental Setting}\label{subsec:setting}

We evaluate our approach in an end-to-end fashion. For a node $c$ in the Chinese BINet, we choose the node $e^*$ which has the highest score as $c$'s counterpart in the English BINet as its translation:

\centerline{
$e^* = \arg\max \limits_{e \in Cand(c)} Score(c,e)$
}

We manually evaluate the quality of translation pairs with top $K$ scores and use \textit{accuracy} as the evaluation metric. Note that a pair $\langle c$,$e \rangle$ is annotated as correct if $e$ is correct translation of $c$ or $c$ is a mention referring to an entity $e$ refers to. The annotation assignment is done by 2 human judges with 92.2\% overlap. The main cause of annotation disagreement is ambiguity of some Chinese words including entities. In the evaluation, we consider $\langle c$,$e\rangle$ correct if both human judges annotate it as correct.

We compare our approach to the following baselines that use various combinations of clues to validate candidates for deciphering the Chinese BINet as well as the state-of-the-art algorithm for language decipherment from non-parallel corpora:


\begin{itemize}
\item Pronunciation validation (\textit{pv}): Use the pronunciation clue only 
\item Translation validation (\textit{tv}): Use the translation clue only 
\item Neighbor validation (\textit{nv}): Use the neighbor clue only to mine translation in a label propagation manner. This baseline is similar to the approach of \cite{tamura2012bilingual}. 
\item Co-burst validation: (\textit{cv}): Use the co-burst clue only
\item \textit{pv+tv} 
\item \textit{pv+tv+nv} 
\item \textit{Bayesian Inference}: The state-of-the-art algorithm \cite{Dou2012} for language decipherment on non-parallel corpora, deciphering a language based on the alignment of its bigram language model to a bigram language model in English using Bayesian inference. In our experimental setting, we first convert BINets to bigram language models: if two nodes $v_1$ and $v_2$ are adjacent in BINets, we treat them as bigrams $v_1,v_2$ and $v_2,v_1$. The count of these bigrams is the weight of the edge between $v_1$ and $v_2$. After normalization, we run the decipherment algorithm of \cite{Dou2012} which outputs the probability distribution of possible candidates given a cipher node (i.e., $P(e|c)$).
\end{itemize} 

For the approaches except Bayesian inference, the score computation function is almost identical to Eq (\ref{eq:overallscore}) except that the weights of the clues which are not used are set to 0.

We used 2009 AFP Chinese/English Gigaword corpora as development set and tuned the parameters by grid search: $\eta=0.25$, $\lambda=0.3$, $\gamma=0.5$, $\delta=0.2$, $\hat{S}_n^{max}=0.4$. The number of iterations in Algorithm \ref{alg:pagerank} is empirically set to 20 which is proven sufficient for convergence of the scores of nodes in the BINets on the development set.

\vspace{-0.1cm}\subsection{Results}
\label{secresults}

\begin{figure}[!t]
\centering
\includegraphics[width=8.5cm]{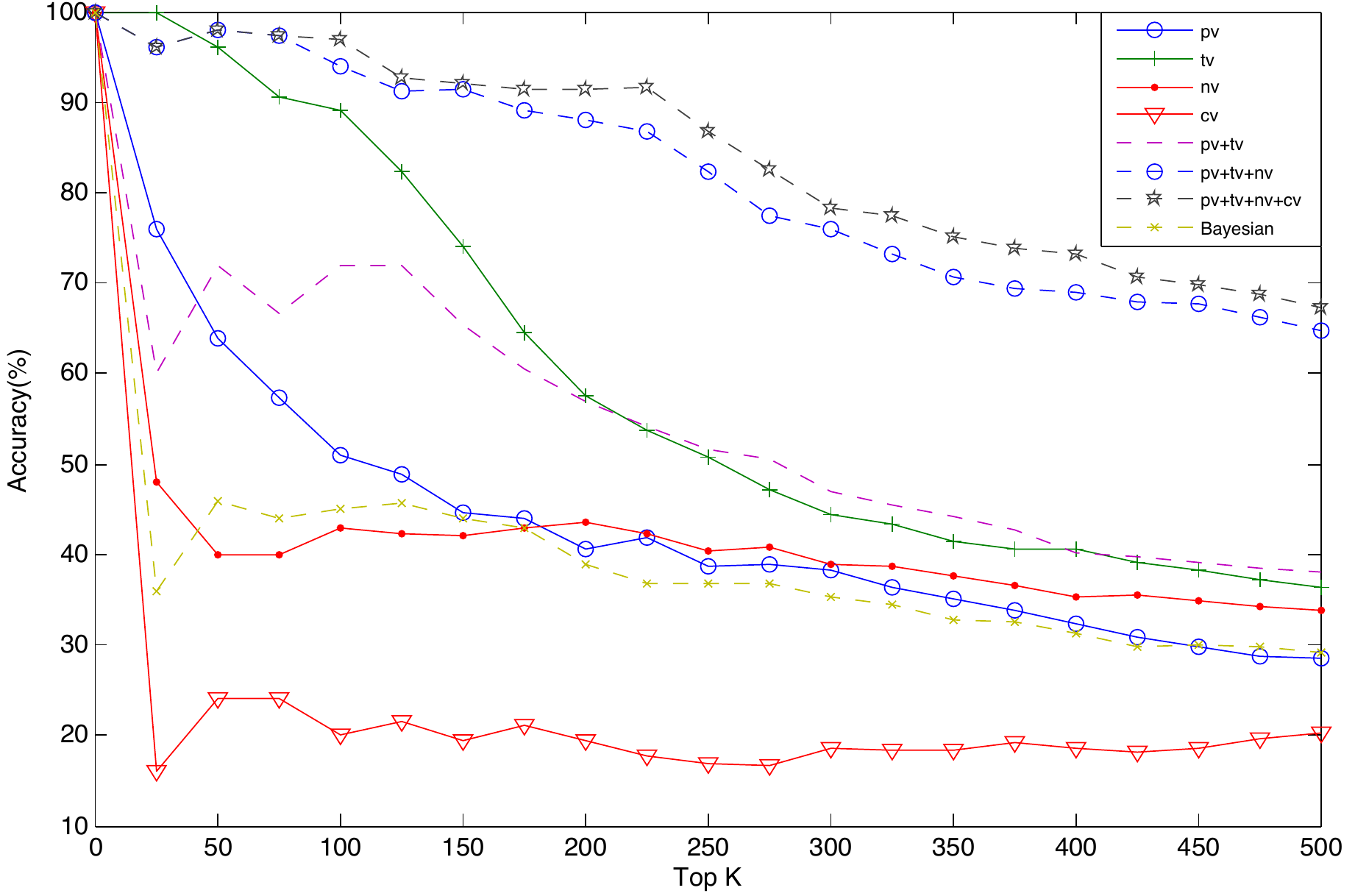}\vspace{-0.2cm}
\caption{Accuracy curves of various approaches. Note that \textit{pv+tv+nv+cv} is our final approach.\label{fig:topk}}\vspace{-0.4cm}
\end{figure}

\begin{table*}[!ht]
\small
\centering
\begin{tabular}{|c|c|c|c|c|c|c|c|c|c|}
\hline
\multirow{2}{*}{} & \multicolumn{3}{c|}{First 6 months} & \multicolumn{3}{c|}{Last 6 months} & \multicolumn{3}{c|}{1 year}  \\ \cline{2-10} 
                  & \#Node     & \#Edge    & \#Doc      & \#Node     & \#Edge    & \#Doc     & \#Node   & \#Edge  & \#Doc   \\ \hline
Chinese                & 3,592      & 17,435    & 8,394      & 3,171      & 12,862    & 8,933     & 7,360    & 33,892  & 17,327  \\ \hline
English                & 5,078      & 28,326    & 114,159    & 2,948      & 43,473    & 72,578    & 8,852    & 85,125  & 186,737 \\ \hline
Time              & \multicolumn{3}{c|}{161.27s}        & \multicolumn{3}{c|}{480.96s}       & \multicolumn{3}{c|}{979.72s} \\ \hline
\end{tabular}
\caption{Run time of the decipherment algorithm on different sizes of streaming data.\label{tab:runtime}}
\end{table*}

We present the results in Figure \ref{fig:topk}. Our approach outperforms all baselines because it considers all available clues for decipherment. Among the baselines, \textit{tv} achieves very high accuracy within top 100 nodes while \textit{pv}'s performance is much lower. The reason is that pronunciation similarity based validation is not suitable for all OOVs, especially common words. For example, the pinyin of 股市(stock market) is ``gushi'', which is very similar to the English word ``gush'' but ``gush'' is not the correct translation of 股市(stock market). It is notable that the accuracy scores of \textit{pv} and \textit{tv} drop dramatically with $K$ increasing because although we can accurately decipher some nodes with high $S_p$ or $S_t$, such nodes are rare. When $S_p$ and $S_t$ are not extremely high, it is risky to decipher a Chinese node just based on $S_p$ or $S_t$. In the dataset we use, only approximately 100 out-of-vocabulary nodes in the Chinese BINet can be effectively deciphered using either \textit{pv} or \textit{tv}, which results in a tumble of accuracy when $K$ increases. \textit{pv+tv} seems to alleviate the problem to some extent: its accuracy does not drop as drastically as \textit{pv} or \textit{tv} because multiple clues allow us to decipher more nodes than a single clue does. However, its accuracy is not desirable -- even lower than \textit{tv} in most cases. Among all the baselines, \textit{cv} seems to perform worst, demonstrating that the co-burst clue alone is far from enough for decipherment.

Compared with \textit{pv}, \textit{tv} and \textit{cv}, \textit{nv} deciphers the nodes in Chinese BINet in a propagation manner but the neighbor clue alone is not sufficient for accurate decipherment. It is notable that the curve of Bayesian inference method is similar to the \textit{nv} because the clues used in these two approaches are almost the same but our graph-based decipherment approach is more flexible to incorporate a variety of clues. When it is combined with \textit{pv+tv}, the performance shows a significant boost and achieves approximately 90\% accuracy in top 200 results though it is slightly inferior to our final approach due to the lack of awareness of co-burst.

\begin{figure}[!t]
\centering
\includegraphics[width=8.2cm]{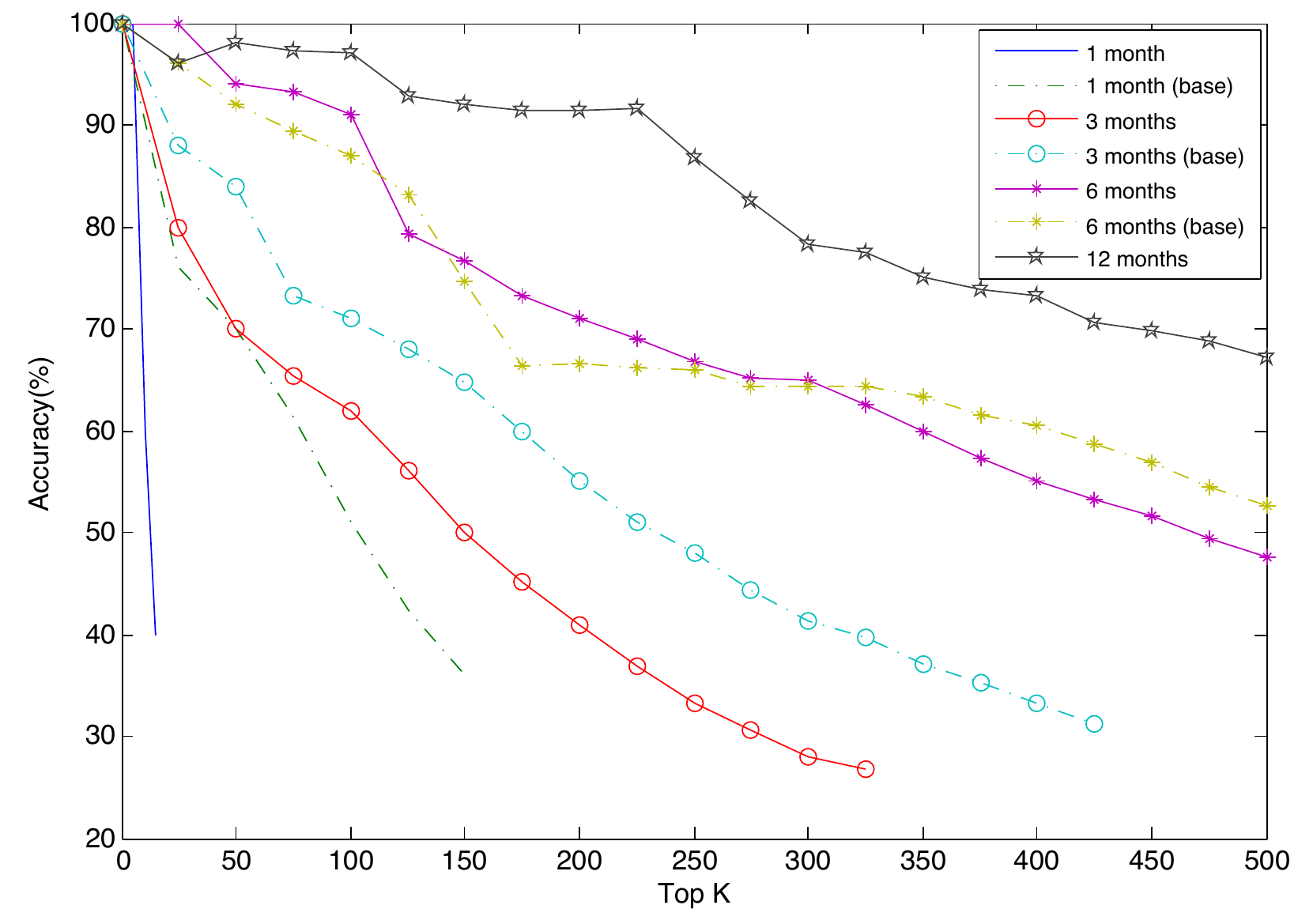}\vspace{-0.2cm}
\caption{Performance of our approach under different sizes of streaming data\label{fig:datasize}. (base) means that the base probability $q_0$ for burst detection is estimated based on the 2009 news streams.}\vspace{-0.35cm}
\end{figure}

We analyzed translation pairs mined by our approach to see how many of them can be obtained by a transliteration model which is often used for name translation. Among top 100 translation pairs, only 9\% can be correctly transliterated by a transliteration model \cite{jiampojamarn2007applying}, demonstrating that our approach can discover large numbers of translation pairs that cannot be transliterated.

Moreover, we investigated the performance of our approach under various sizes of data provided, as shown in Figure \ref{fig:datasize}. As observed, when the data size is too small, the approach works poorly. One reason is that there are few nodes to be deciphered in BINets that can be aligned; the other reason is that the burst detection algorithm cannot work well in a small time frame because the base probability $q_0$ cannot be estimated precisely. For example, an important event (e.g., the World Cup) that is frequently mentioned within a month may not be recognized as a bursty event if only the data in this month is available. When we estimate the base probability $q_0$ based on a larger data stream (2009 news streams), the performance of our approach on 1-month data and 3-month data is significantly improved while the performance on 6-month data is not, demonstrating that 6-month data is sufficient for base probability estimation.

It is clearly seen from Figure \ref{fig:datasize} that the more data we have, the more translation pairs that can be mined. Considering massive streaming data generated every day, if the approach can be applied to the data streams, it is possible to monitor the streaming data and automatically mine the translation of new words or entities in a never-ending fashion.

For the sake of that goal, a main requirement for our approach is that the algorithm should be efficient or can be applied in parallel. To verify the efficiency of our approach, we split the streaming data into two parts based on timestamps: first 6 months and last 6 months, and tested the approach on these two parts.

Table \ref{tab:runtime} shows the run time of the decipherment algorithm on different sizes of streaming data, which is measured on a workstation with Intel Xeon 3.5 GHz CPU and 64GB RAM. As data increases, it will take longer time for decipherment. However, it is notable that the decipherment processes for the data of the first 6 months and the last 6 months are independent and thus it is possible to run the decipherment algorithm on these two datasets in parallel and merge the decipherment results.

\begin{figure}[!t]
\centering
\includegraphics[width=7.5cm]{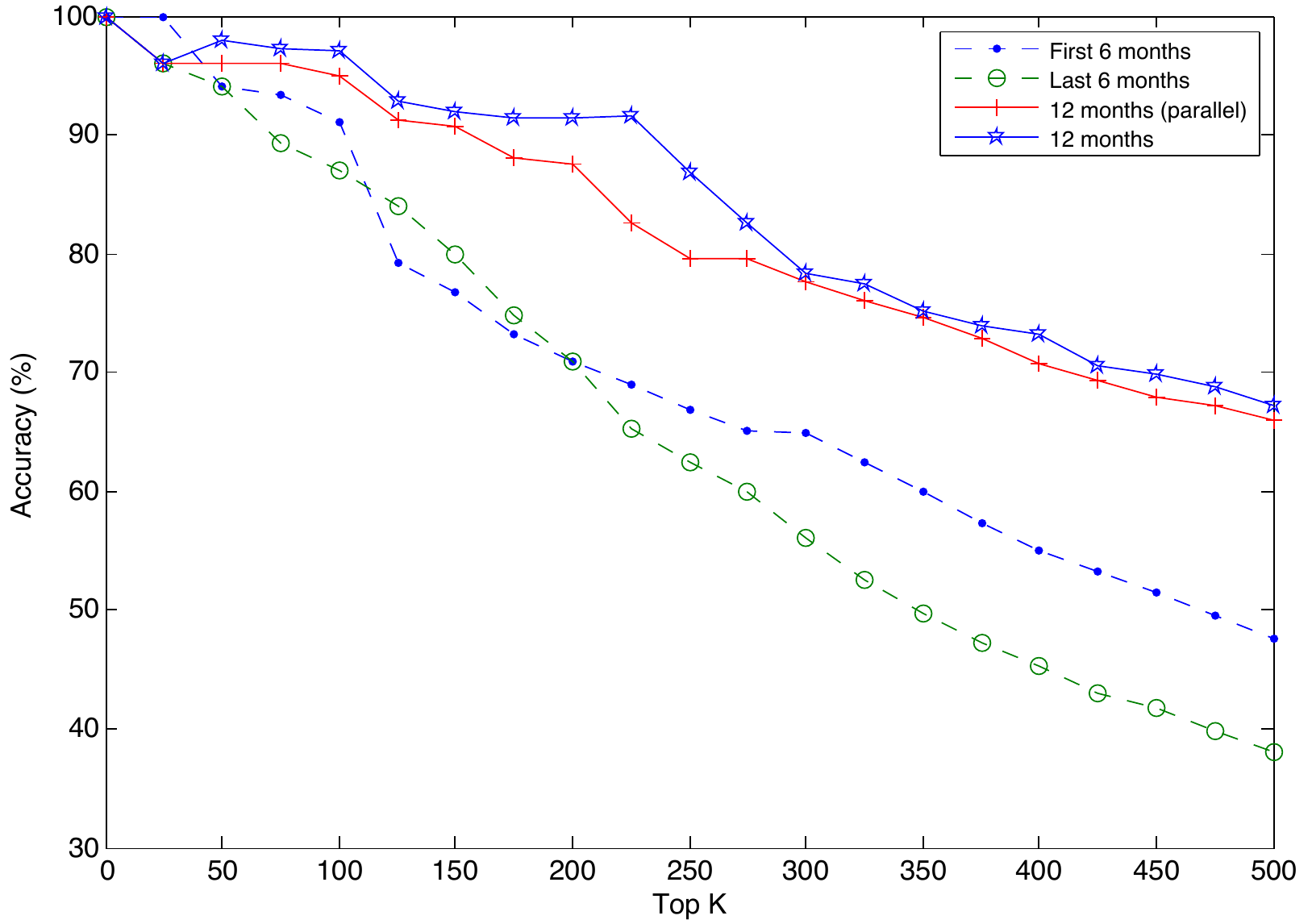}\vspace{-0.2cm}
\caption{Performance of decipherment algorithm in parallel\label{fig:online}}\vspace{-0.3cm}
\end{figure}

Figure \ref{fig:online} shows the performance of the decipherment algorithm in the parallel fashion. We can see that deciphering in parallel does not result in a significant decrease of accuracy. Therefore, we can split a text stream into several small parts and decipher them in parallel, which makes the decipherment more efficient. For the example in Table \ref{tab:runtime}, if we decipher the text streams of the first 6 months and the last 6 months in parallel, the run time of decipherment would be 480.96 seconds assuming the time for merging the results is negligible.

\begin{figure}[!t]
\centering
\includegraphics[width=7.5cm]{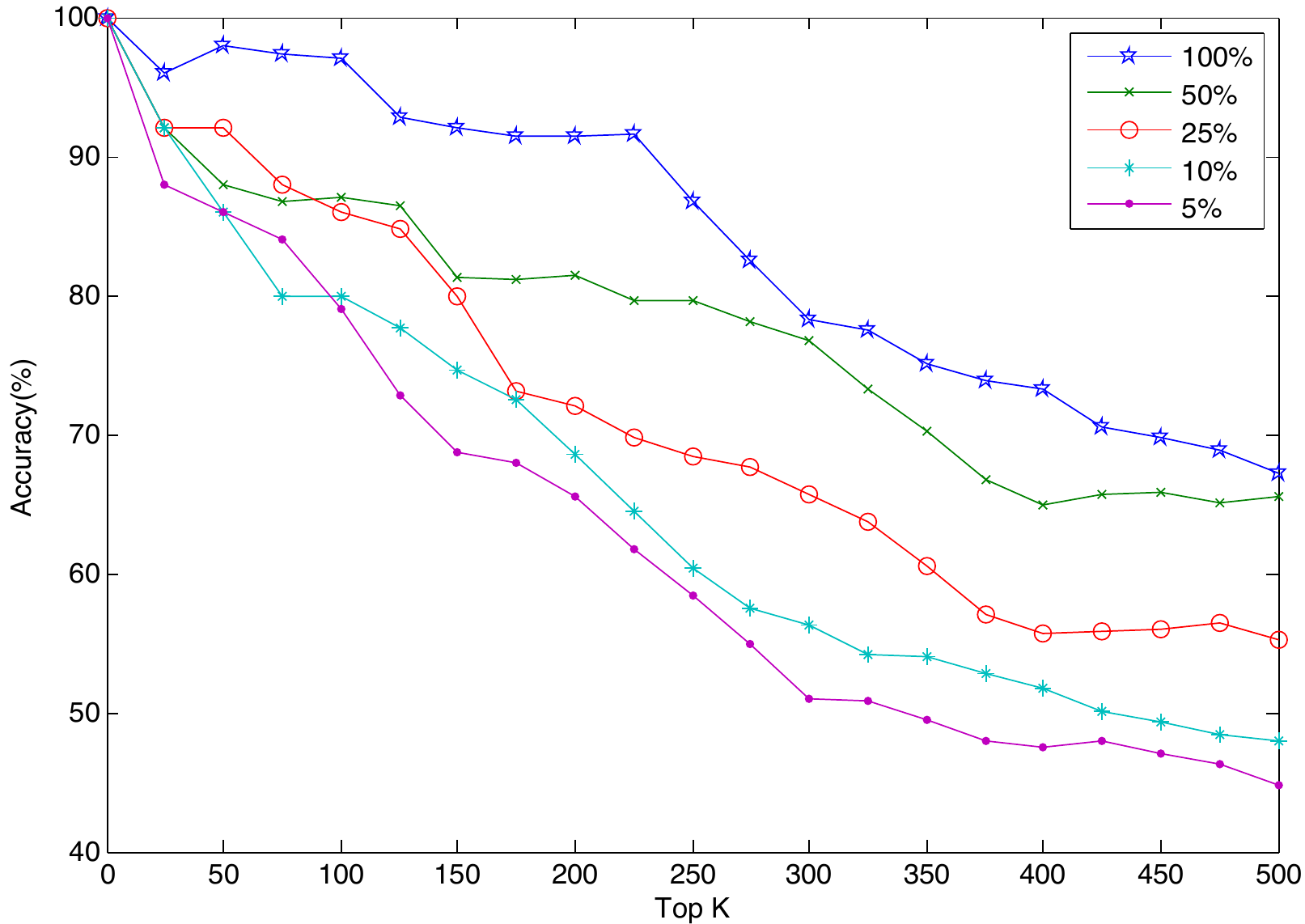}\vspace{-0.2cm}
\caption{Accuracy curves of our approach with different sizes of bi-lingual lexicon.\label{fig:scurve}\vspace{-0.2cm}}
\end{figure}

We also study the effect of the size of the bi-lingual lexicon on the performance. We randomly sample different sizes of entries from the original bi-lingual lexicon as new bi-lingual lexicons. The results are shown in Figure \ref{fig:scurve}. 
We can see that the decipherment accuracy increases as the size of bi-lingual lexicon increases because more prior knowledge can help better decipher the BINet.

\begin{figure}[t]
\centering
\includegraphics[width=7.5cm]{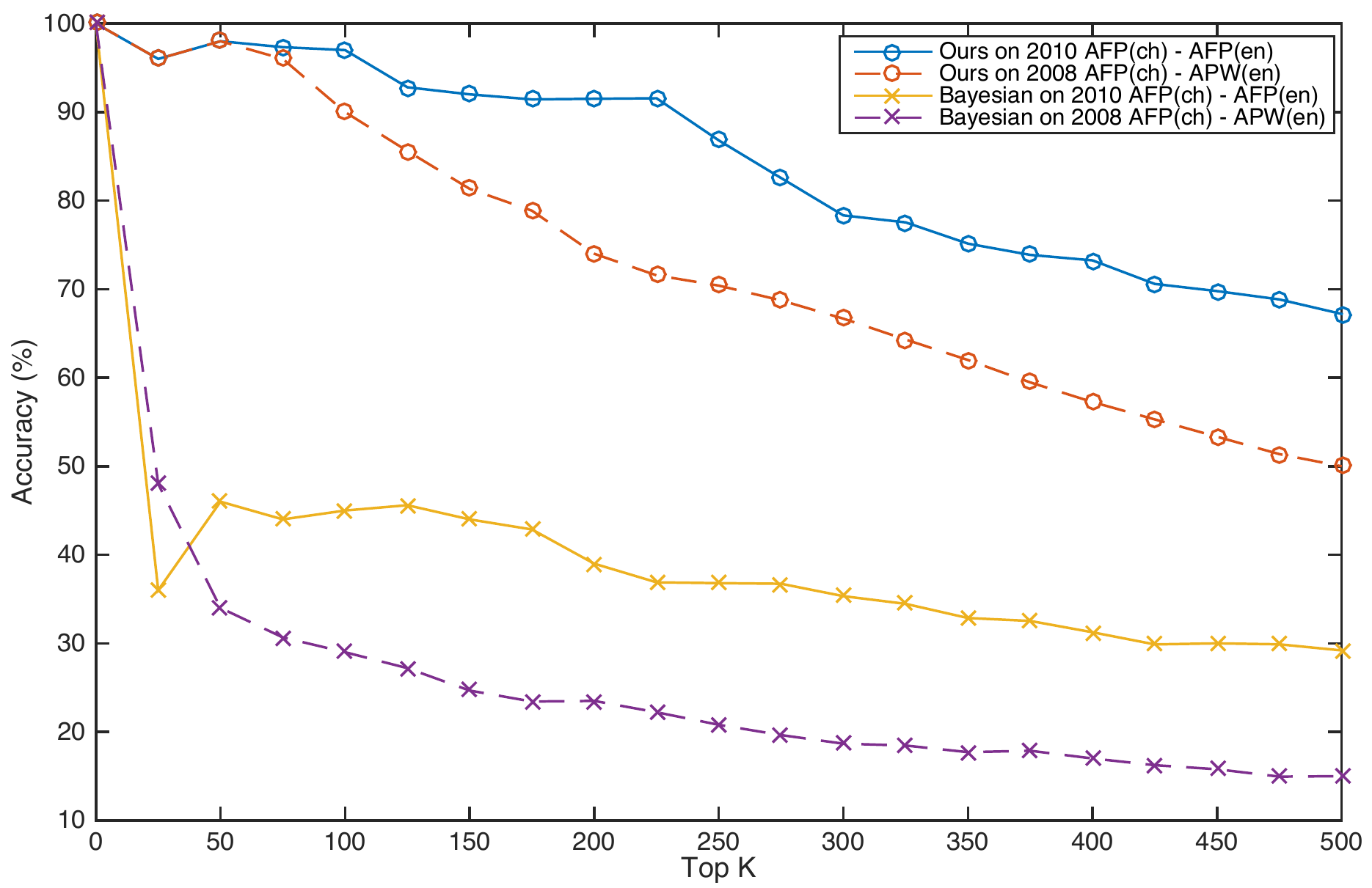}\vspace{-0.2cm}
\caption{Decipherment performance on multiple datasets.\label{fig:multi_data}}\vspace{-0.3cm}
\end{figure}
\begin{table}[t]
\small
\centering
\begin{tabular}{ccc}
\hline
\bf Model & \bf $Acc_1$(2010) & \bf $Acc_1$(2008) \\ \hline
\textsc{Context} & 0.32\% & 0.37\%\\ 
\textsc{Diverse} & 5.22\% & 4.25\%\\
\textsc{CoLP} & 0.32\% & 0.46\%\\
\textsc{SimLP} & 0.49\% & 0.46\%\\
\textsc{Bayesian(LM)} & 0.57\% & 0.55\%\\
\textsc{Bayesian(BINet)} & 11.17\% & 4.81\% \\
Ours & \bf 28.14\% & \bf 19.50\%\\
\hline
\end{tabular}\vspace{-0.1cm}
\caption{Performance of translating bursty words.\label{tab:biset}}\vspace{-0.5cm}
\end{table}

In order to test the generalization ability, we evaluate our decipherment approach using the same parameters on another coordinated text streams -- AFP Chinese and APW English news stream in 2008. The result is shown in Figure \ref{fig:multi_data}. Our decipherment approach consistently outperforms the Bayesian inference approach and still deciphers top 100 nodes in very high accuracy even though the curve of the streams in 2008 is lower than those in 2010. The difference of the performance on different datasets mainly arises from the differences of topic overlaps. In the streams of 2010, the Chinese and English news streams are from the same news agency (i.e., AFP). Therefore, the topic overlaps between the Chinese and English streams are much more than those of the 2008 coordinated streams, which allows more nodes in the Chinese BINet to be deciphered correctly.

\begin{table*}[!t]
\centering
\small
\scalebox{0.95}
{
\begin{tabular}{ccccc}
\hline
\bf Node & {\bf Chinese (burst period)}     & {\bf English (burst period)}             & {\bf Score} & \bf Main clue \\ \hline
 1  & 贝鲁斯柯尼 (346-348) & Berlusconi (344-348) &  0.990       & pronunciation \\
   2 & 佛州 (249-254) & Florida (249-254) & 0.632 & translation \\
  3  & 卫报 (332-337) & Guardian (334-336) & 0.407 & translation\\  
  4  & 曼城 (360-363)      & Manchester City (358-361) & 0.579  & translation  \\
  5  & 鸠山 (147-158)      & Hatoyama (147-158)        & 0.474  & neighbor  \\
 6  & 千岛群岛 (304-306)    & Kuril Islands (302-307)   & 0.121 & neighbor \\ 
 7   & 国际奥林匹克委员会 (38-45) & IOC (36-44)               & 0.427 & neighbor\\
 8  & 恋童癖 (102-104)     & paedophilia (90-108)      & 0.090  &neighbor \\
 9 & 小威 (147-153)      & Serena Williams (146-148) & 0.152 & neighbor \\ 
 10   & 东协 (200-203)      & ASEAN (198-203)           & 0.598  & translation    \\ 
 11  & 东南亚国家协会 (299-302) & ASEAN (299-303)           & 0.522  & translation      \\ 
 12 & 沉舰 (137-148)      & Cheonan (138-144)         & 0.606 & neighbor               \\
 13   & 天安舰 (137-140)     & Cheonan (138-144)         & 0.350  & neighbor    \\
 14    & 央行 (129-130) & European Central Bank   (125-129) & 0.262 & translation \\
 15 & 央行 (75-76) & Bank of Japan (73-76)   & 0.096 & neighbor   \\ 
  16  & 翁山 (310-311)      & Suu Kyi (308-315)         & 0.434  & neighbor \\
 17  & 苏姬 (310-311)     & Suu Kyi (308-315)        & 0.434   & neighbor \\\hline 
\end{tabular}
}\vspace{-0.1cm}
\caption{Examples of decipherment results.\label{tab:dresult}}\vspace{-0.35cm}
\end{table*}

Also, we evaluated our approach on extracting translation of bursty words which are the most important information in the text stream by comparing conventional bilingual lexicon extraction methods. Specifically, we test how many OOV words appearing in our BINet are correctly translated. In total, there are 1,226 and 1,082  distinct OOV words (excluding incorrectly segmented words) in our Chinese BINets of the 2010 and 2008 streams respectively. For the BINet-based approaches, we find a Chinese word $w$'s English translation $w^*$ as follows:

\centerline{\small $w^*=w(e^*)$
}
\centerline{\small $e^*=arg\max_{e \in V_e}\max_{c \in V_c(w)}Score(c,e)$
}

\noindent where $V_c(w)$ is set of Chinese nodes whose word is $w$, and $w(e)$ denotes the word of node $e$.
For the other bilingual lexicon extraction models, we use the same datasets as comparable corpora and the same bilingual lexicon as seeds to extract translation of the OOV words. Accuracy is used to measure the proportion of the words that are correctly translated, as \cite{tamura2012bilingual} did.

Table \ref{tab:biset} shows the performance of our approach and typical bilingual lexicon extraction methods. \textsc{Context} is proposed by \cite{Fung1998} which extracts translation pairs based on context similarity of words. \textsc{Diverse} is proposed by \cite{schafer2002inducing}, which is a variant of \cite{Fung1998} by adding pinyin and temporal similarity. \textsc{CoLP} and \textsc{SimLP} are label propagation methods proposed by \cite{tamura2012bilingual} on word co-occurrence and similarity graph respectively. \textsc{Bayesian} is a language decipherment model by \cite{Dou2012}. Specially, we evaluate it in two settings (i.e., based on traditional language models and BINets) because it can be run on the BINet as discussed in Section \ref{subsec:setting}. According to Table \ref{tab:biset}, our approach significantly outperforms the other approaches on both datasets due to the rich clues and our candidate generation strategy that can largely narrow down the number of candidates, showing its advantages for translating bursty words in text streams. It is also notable that the BINet-based \textsc{Bayesian} improves the LM-based counterpart significantly, demonstrating the effectiveness of BINets on bi-lingual lexicon extraction from coordinated streams.

Finally, we list some representative nodes deciphered by our approach in Table \ref{tab:dresult}. 
Once the BINets are constructed and deciphered, many other powerful knowledge extraction mechanisms are enabled. For example, a natural by-product of our approach is detecting the influential roles of each community. Moreover, we can utilize the BINets to disambiguate entities, construct their profiles and predict hidden relations simultaneously during the decipherment. For example, 
for node 9 in Table \ref{tab:dresult}, deciphering the nickname ``小威'' into Serena Williams can greatly benefit cross-lingual entity linking. Nodes 10-11 and 12-13 also demonstrate that the potential effect on entity linking and coreference resolution. Nodes 14-15 also show that the deciphered BINets can help entity disambiguation. ``央行(Central bank)'' may refer to different entities during different burst periods. Moreover, the deciphered BINets can also improve entity extraction. 
For example, in nodes 16-17, 翁山苏姬(Aung San Suu Kyi) is not recognized as a person name by the Chinese name tagger; instead, it is mistakenly separated into two words -- 翁山(Aung San) and 苏姬(Suu Kyi). However, since 翁山(Aung San) and 苏姬(Suu Kyi) are deciphered into the same English named entity -- Suu Kyi, we can merge them back to form the correct entity.

\vspace{-0.25cm}\section{Related Work}

Burst patterns have been studied for decades in data mining and natural language processing communities. \cite{swan2000timemines,kleinberg2003bursty,yao2010temporal,tamura2012location,zhao2012novel} studied burst detection problem from text streams. \cite{swan2000timemines,fung2005parameter,Wang2007,he2007using,wang2009mining,diao2012finding,zhao2012novel,yan2015probabilistic} employed burst information to find bursty topics and events.

Extensive research has also been done on bilingual lexicon induction (e.g.,~\cite{Fung1998,Rapp1999,koehn2002learning,schafer2002inducing,Shao2004,schafer2006translation,Hassan2007,Haghighi2008,Udupa2009,Klementiev2010,tamura2012bilingual,Irvine2013,irvine2015discriminative,kiela2015visual}) and name translation mining (e.g.,~\cite{sproat2006named,klementiev2006weakly,udupa2008mining,ji2009mining,You2010,kotov2011mining,Lin2011,sellami2014mining}) from non-parallel corpora. Some of these approaches also exploited temporal similarity and context information. We proposed new and richer clues including burst and graph topological structure. The nodes in BINet are not limited to named entities. More importantly, it is much cheaper to construct BINets than traditional information networks which usually rely on supervised information extraction and large numbers of language-specific resources.

In contrast to previous cross-lingual projection work like data transfer~\cite{Pado2009} and model transfer~\cite{McDonald2011}, we do not require any parallel data. 
Our BINet construction method was inspired by~\cite{zhao2012novel,sayyadi2013graph}. Our work is the first to apply it to a cross-lingual setting. This is also the first attempt to apply the decipherment idea (e.g., \cite{Ravi2011,Dou2012,Dou2014}) to graph structures instead of sequence data. Another work related to ours is \cite{sproat2006named,kotov2011mining}, which used phonetic transliteration and frequency correlation to discover transliteration of entities. Compared with their work, this paper addresses a more general problem -- we not only focus on named entities but all kinds of out-of-vocabulary words and phrases, which means that the number of candidates of a given node is much larger. Moreover, besides the named entity transliteration pairs, the BINets combined with a variety of clues (i.e., similarities of pronunciation, translation, context and temporal correlation) can discover word/phrase and named entity translation pairs that cannot be discovered by their work.

\vspace{-0.25cm}\section{Conclusions and Future Work}
In this paper, we propose an approach to construct and decipher burst information networks constructed from foreign languages, as a novel and unique way to align streaming data. For the first time we propose to model the translation pair mining from non-parallel corpora as a network decipherment problem. Based on the characteristics of co-burst across languages, we proposed 4 novel and effective clues for decipherment. Our approach is language-independent, efficient, effective and can be easily implemented, thus it is useful for never-ending language knowledge acquisition, machine translation and cross-lingual information retrieval systems.

In the future, we plan to introduce richer graph-based features (e.g., meta-paths \cite{Han2010}). This self-boosting framework can also mutually enhance extraction and translation acquisition, and demonstrate a successful joint inference procedure across disparate knowledge sources from multiple languages.




\bibliographystyle{unsrt}
\bibliography{CrossKN_short}

\begin{thebibliography}{10}

\bibitem{Kedzie2015}
A.~Klementiev and C.~Callison-Burch.
\newblock Predicting salient updates for disaster summarization.
\newblock In {\em ACL}, 2015.

\bibitem{Wang2007}
X.~Wang~et al.
\newblock Mining correlated bursty topic patterns from coordinated text
  streams.
\newblock In {\em KDD}, 2007.

\bibitem{zheng2012cross}
S.H. Zheng~et al.
\newblock Cross-lingual topic alignment in time series japanese/chinese news.
\newblock In {\em PACLIC}, 2012.

\bibitem{Han2010}
J.~Han~et al.
\newblock Mining heterogeneous information networks.
\newblock In {\em KDD}, 2010.

\bibitem{Li2014}
Q.~Li et~al.
\newblock Constructing information networks using one single model.
\newblock In {\em EMNLP}, 2014.

\bibitem{kleinberg2003bursty}
J.~Kleinberg.
\newblock Bursty and hierarchical structure in streams.
\newblock {\em Data Mining and Knowledge Discovery}, 2003.

\bibitem{zhao2012novel}
X.~Zhao~et al.
\newblock A novel burst-based text representation model for scalable event
  detection.
\newblock In {\em ACL}, 2012.

\bibitem{sayyadi2013graph}
H.~Sayyadi and L.~Raschid.
\newblock A graph analytical approach for topic detection.
\newblock {\em ACM Transactions on Internet Technology (TOIT)}, 2013.

\bibitem{koehn2002learning}
P.~Koehn and K.~Knight.
\newblock Learning a translation lexicon from monolingual corpora.
\newblock In {\em ACL workshop on Unsupervised lexical acquisition}, 2002.

\bibitem{Irvine2013}
A.~Irvine and C.~Callison-Burch.
\newblock Supervised bilingual lexicon induction with multiple monolingual
  signals.
\newblock In {\em NAACL}, 2013.

\bibitem{Ravi2011}
S.~Ravi~et al. and K.~Knight.
\newblock Deciphering foreign language.
\newblock In {\em ACL}, 2011.

\bibitem{Dou2012}
Q.~Dou~et al.
\newblock Large scale decipherment for out-of-domain machine translation.
\newblock In {\em EMNLP}, 2012.

\bibitem{Dou2014}
Q.~Dou~et al.
\newblock Beyond parallel data: Joint word alignment and decipherment improves
  machine translation.
\newblock In {\em EMNLP}, 2014.

\bibitem{ji2009mining}
H.~Ji.
\newblock Mining name translations from comparable corpora by creating
  bilingual information networks.
\newblock In {\em the 2nd Workshop on Building and Using Comparable Corpora:
  from Parallel to Non-parallel Corpora}, 2009.

\bibitem{Lin2011}
W.~Lin~et al.
\newblock Unsupervised language-independent name translation mining from
  wikipedia infoboxes.
\newblock In {\em EMNLP Workshop on Unsupervised Learning for NLP}, 2011.

\bibitem{graff2005chinese}
D.~Graff and K.~Chen.
\newblock Chinese gigaword.
\newblock {\em LDC Catalog No.: LDC2003T09, ISBN}, 1:58563--58230, 2005.

\bibitem{graff2003english}
D.~Graff~et al.
\newblock English gigaword.
\newblock {\em Linguistic Data Consortium, Philadelphia}, 2003.

\bibitem{manning-EtAl:2014:P14-5}
C.D. et~al. Manning.
\newblock The {Stanford} {CoreNLP} natural language ssing toolkit, 2014.

\bibitem{zens2004improvements}
R.~Zens and H.~Ney.
\newblock Improvements in phrase-based statistical machine translation.
\newblock In {\em HLT-NAACL}, 2004.

\bibitem{tamura2012bilingual}
A.~Tamura~et al.
\newblock Bilingual lexicon extraction from comparable corpora using label
  propagation.
\newblock In {\em EMNLP}, 2012.

\bibitem{jiampojamarn2007applying}
S.~Jiampojamarn~et al.
\newblock Applying many-to-many alignments and hidden markov models to
  letter-to-phoneme conversion.
\newblock In {\em NAACL}, 2007.

\bibitem{Fung1998}
P.~Fung and L.Y. Yee.
\newblock An ir approach for translating new words from nonparallel and
  comparable texts.
\newblock In {\em COLING-ACL}, 1998.

\bibitem{schafer2002inducing}
C.~Schafer and D.~Yarowsky.
\newblock Inducing translation lexicons via diverse similarity measures and
  bridge languages.
\newblock In {\em CoNLL}, 2002.

\bibitem{swan2000timemines}
R.~Swan and D.~Jensen.
\newblock Timemines: Constructing timelines with statistical models of word
  usage.
\newblock In {\em KDD Workshop on Text Mining}, 2000.

\bibitem{yao2010temporal}
J.~Yao~et al.
\newblock Temporal and social context based burst detection from folksonomies.
\newblock In {\em AAAI}, 2010.

\bibitem{tamura2012location}
K.~Tamura and H.~Kitakami.
\newblock Location-based burst detection algorithm in spatiotemporal document
  stream.
\newblock In {\em DMIN}, 2012.

\bibitem{fung2005parameter}
G.P.C. Fung~et al.
\newblock Parameter free bursty events detection in text streams.
\newblock In {\em VLDB}, 2005.

\bibitem{he2007using}
Q.~He~et al.
\newblock Using burstiness to improve clustering of topics in news streams.
\newblock In {\em ICDM}, 2007.

\bibitem{wang2009mining}
X.~Wang~et al.
\newblock Mining common topics from multiple asynchronous text streams.
\newblock In {\em WSDM}, 2009.

\bibitem{diao2012finding}
Q.~Diao~et al.
\newblock Finding bursty topics from microblogs.
\newblock In {\em ACL}, 2012.

\bibitem{yan2015probabilistic}
X.~Yan~et al.
\newblock A probabilistic model for bursty topic discovery in microblogs.
\newblock In {\em AAAI}, 2015.

\bibitem{Rapp1999}
R.~Rapp.
\newblock Automatic identification of word translations from unrelated english
  and german corpora.
\newblock In {\em ACL}, 1999.

\bibitem{Shao2004}
L.~Shao and H.T. Ng.
\newblock Mining new word translations from comparable corpora.
\newblock In {\em COLING}, 2004.

\bibitem{schafer2006translation}
Charles~F Schafer~III.
\newblock {\em Translation discovery using diverse similarity measures}.
\newblock Johns Hopkins University, 2006.

\bibitem{Hassan2007}
A.~Hassan~et al.
\newblock Improving named entity translation by exploiting comparable and
  parallel corpora.
\newblock In {\em RANLP}, 2007.

\bibitem{Haghighi2008}
A.~Haghighi~et al.
\newblock Learning bilingual lexicons from monolingual corpora.
\newblock In {\em ACL}, 2008.

\bibitem{Udupa2009}
R.~Udupa~et al.
\newblock Mint: A method for effective and scalable mining of named entity
  transliterations from large comparable corpora.
\newblock In {\em EACL}, 2009.

\bibitem{Klementiev2010}
A.~Klementiev and C.~Callison-Burch.
\newblock Bilingual lexicon induction for low-resource languages.
\newblock In {\em JHU Technical Report}, 2010.

\bibitem{irvine2015discriminative}
A.~Irvine and C.~Callison-Burch.
\newblock Discriminative bilingual lexicon induction.
\newblock {\em Computational Linguistics}, 2015.

\bibitem{kiela2015visual}
D.~Kiela~et al.
\newblock Visual bilingual lexicon induction with transferred convnet features.
\newblock In {\em EMNLP}, 2015.

\bibitem{sproat2006named}
R.~Sproat~et al.
\newblock Named entity transliteration with comparable corpora.
\newblock In {\em ACL}, 2006.

\bibitem{klementiev2006weakly}
A.~Klementiev and D.~Roth.
\newblock Weakly supervised named entity transliteration and discovery from
  multilingual comparable corpora.
\newblock In {\em COLING-ACL}, 2006.

\bibitem{udupa2008mining}
R.~Udupa~et al.
\newblock Mining named entity transliteration equivalents from comparable
  corpora.
\newblock In {\em CIKM}, 2008.

\bibitem{You2010}
G.~You et~al.
\newblock Mining name translations from entity graph mapping.
\newblock In {\em EMNLP}, 2010.

\bibitem{kotov2011mining}
A.~Kotov~et al.
\newblock Mining named entities with temporally correlated bursts from
  multilingual web news streams.
\newblock In {\em WSDM}, 2011.

\bibitem{sellami2014mining}
R.~Sellami~et al.
\newblock Mining named entity translation from non parallel corpora.
\newblock In {\em FLAIRS}, 2014.

\bibitem{Pado2009}
S.~Pado and M.~Lapata.
\newblock Cross-lingual annotation projection for semantic roles.
\newblock {\em Journal of Artificial Intelligence Research}, 36, 2009.

\bibitem{McDonald2011}
R.~McDonald et~al.
\newblock Multi-source transfer of delexicalized dependency parsers.
\newblock In {\em EMNLP}, 2011.

\end{thebibliography}
\end{CJK}
\end{document}